\documentclass{article} % For LaTeX2e
\usepackage{iclr2027_conference,times}

\usepackage{amsmath,amsfonts,bm}

\def\eqref#1{equation~\ref{#1}}
\def\1{\bm{1}}

\DeclareMathAlphabet{\mathsfit}{\encodingdefault}{\sfdefault}{m}{sl}
\SetMathAlphabet{\mathsfit}{bold}{\encodingdefault}{\sfdefault}{bx}{n}

\usepackage{hyperref}
\usepackage{url}
\usepackage{xspace}
\usepackage{graphicx}
\usepackage{subcaption}
\usepackage{float}
\usepackage{booktabs}
\usepackage{makecell}

\title{FastCI: Efficient GPU-Intensive CI for LLM Training Frameworks}

\author{
\rule{0pt}{48pt}% Add space between the title and the author list.
\makebox[\dimexpr\textwidth-2\tabcolsep\relax][c]{%
\begin{tabular}{@{}c@{}}
\textbf{Tianshuo Qiao$^{1}$, Naiqian Zheng$^{1}$, Xiaopeng Liu$^{2}$, Shuguang Wang$^{2}$,} \\
\textbf{Diandian Gu$^{2}$, Xuanzhe Liu$^{1}$, Xin Jin$^{1,}$\thanks{Corresponding author.}} \\
{\normalfont $^{1}$Peking University \quad $^{2}$ByteDance Seed}
\end{tabular}%
}
}

\newcommand{\sysname}{FastCI\xspace}
\newcommand{\company}{ByteDance\xspace}

\iclrfinalcopy % Uncomment for camera-ready version, but NOT for submission.
\begin{document}

\maketitle
\lhead{Under review at ICLR 2027}

\begin{abstract}
As large language models (LLMs) keep growing in size and complexity, their training frameworks
evolve at a rapid pace as well. Therefore, continuous integration (CI) is critical for maintaining the
quality and stability of these frameworks. However, unlike traditional software, CI for LLM training
frameworks relies on GPU-intensive tests, which usually involve complete model training or
evaluation. This leads CI itself to become a new bottleneck for fast-paced development.
In this paper, we introduce \sysname, a framework that improves the efficiency of CI for LLM training
frameworks.
\sysname leverages runtime evidence to select
affected tests and prune tests that execute changed code in equivalent contexts. Then \sysname prioritizes
high-risk tests to expose potential failures earlier, and optimizes test workloads along dimensions
outside the intended validation scope of each test. Evaluated on the CI workload of our LLM training
framework, \sysname reduces the CI latency by 77.5\% and the GPU resource usage by 63.9\%, while
improving the modified code coverage retention by 3.2\%, compared with the currently deployed CI pipelines. \sysname has now
been integrated into the CI pipelines of our LLM training framework at \company.
\end{abstract}

\section{Introduction}
\label{sec:introduction}

As large language models (LLMs) keep growing in size and complexity and are adopted in a wider range of tasks
\citep{vaswani2017attention,brown2020language,hoffmann2022training}, their training frameworks have evolved from simple
libraries to sophisticated systems. For example, training frameworks may combine data, tensor, and pipeline parallelism with
partitioned optimizer states in large scale training, or jointly manage vision and language components in multimodal training
\citep{narayanan2021megatron,rajbhandari2020zero,alayrac2022flamingo,liu2023llava}.
To preserve the quality and stability
of these frameworks, continuous integration (CI) has become a critical part of the development process.
CI refers to continuously building and testing code changes, commonly when a merge request (MR) is submitted or updated.
Unlike traditional software, whose CI workloads are typically dominated by unit tests, LLM training frameworks rely heavily on GPU-intensive end-to-end
tests, since many behaviors, including successful training completion and numerical alignment, can
be verified only through complete model training or evaluation. Validating these behaviors requires
an extensive test suite that takes hours and consumes substantial GPU resources, making CI a new bottleneck
in fast-paced LLM framework development.

In production CI pipelines, long latency and high GPU resource consumption are even more severe \citep{hilton2016usage,gandiva2018,tiresias2019,memon2017taming}. The master branch of our training framework receives more than
200 MRs per week, and each MR may trigger CI pipelines multiple times as developers push commits to refine code or fix failures. Coding agents have further increased the frequency of code changes, accelerating development while also adding to the CI burden~\citep{malhotra2026agenticci}.
To handle this workload, we maintain a dedicated CI resource pool comprising
thousands of GPUs. Nevertheless, this resource pool remains saturated, and a CI run takes more than four hours on average.
% Since an MR may trigger multiple runs as it evolves, developers experience this delay repeatedly.
This problem stems not only from the high cost of individual
end-to-end tests, but also from inefficiencies in how existing CI pipelines validate the code changes.
In fact, many tests are unnecessary, as they do not cover the changed code.

\begin{figure}[t]
\centering
\vspace{-2em}
\includegraphics[width=\textwidth]{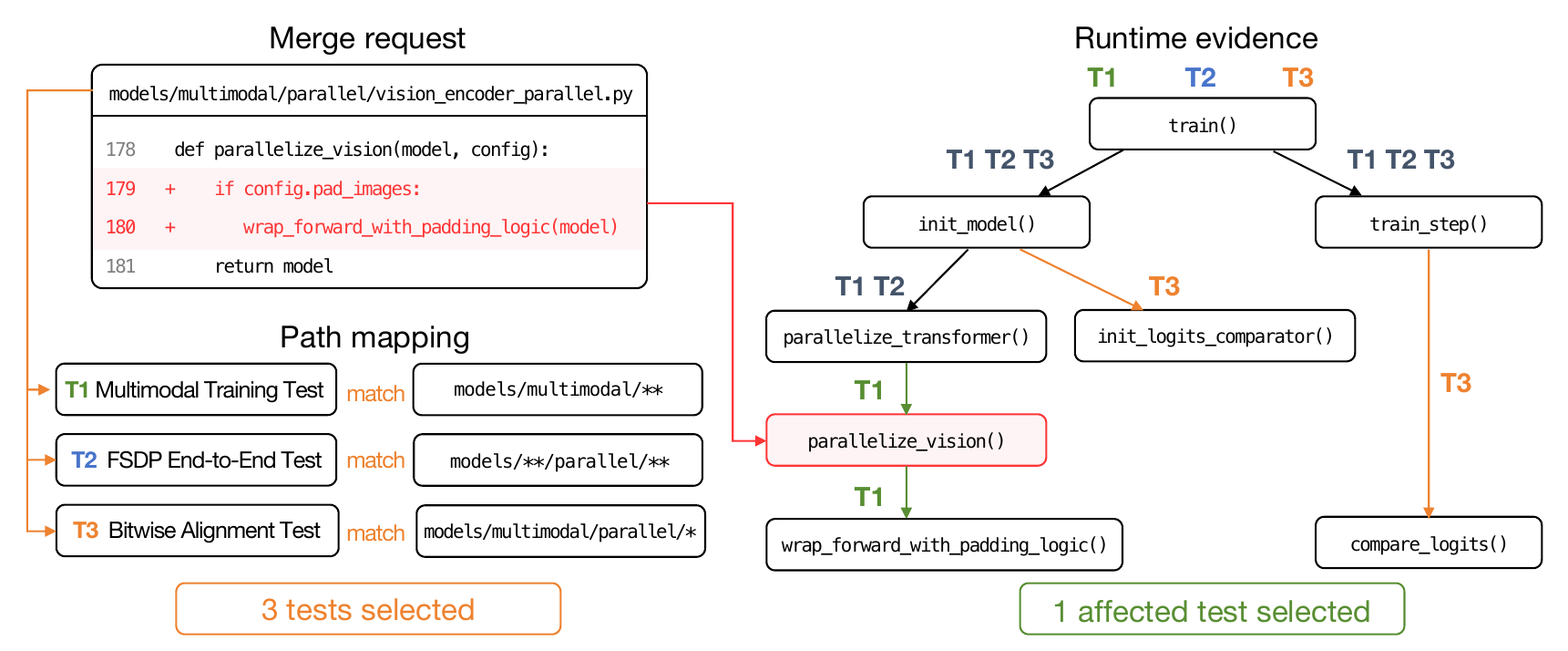}
\caption{Test selection using path mappings and runtime evidence.}
\label{fig:intro:motivation}
\vspace{-1.05em}
\end{figure}

To prevent CI from triggering full test suites for each MR, path-based test selection
has been adopted in open source projects and industry practice \citep{luu2026vllmci,rayproject2026testselection}. This approach selects tests based on
the mapping from files to tests, which is built and maintained manually in our deployed CI pipelines. There are two limitations
preventing this approach from addressing the CI bottleneck for LLM training frameworks effectively. First,
the mapping from files to tests is too coarse-grained to capture the dependencies between code changes and tests
precisely, still causing many unnecessary tests to be executed. Second, maintaining the mapping manually is
error-prone as the codebase evolves rapidly, and incomplete mappings can lead to false negatives in test selection.
Figure~\ref{fig:intro:motivation} illustrates over-selection by path-based test selection. In this example, an MR modifies the function
\texttt{parallelize\_vision} within \texttt{vision\_encoder\_parallel.py}. Path-based test selection includes all
tests that depend on the modified file, while runtime evidence shows that only the Multimodal Training Test
executes the modified function. This suggests that precise test selection should move from static file-level path mapping
to function-level runtime evidence.

Beyond test selection, how the selected tests are executed presents further optimization opportunities \citep{li2024testboost,dutta2021tera}. In our CI pipelines,
tests follow first-in-first-out scheduling, which may take a long time to expose CI failures. Exposing failures earlier allows
developers to manually stop the CI run and fix them sooner, saving time and GPU resources. In addition,
end-to-end tests usually reuse the same workloads whose model sizes and numbers of training steps exceed what is necessary for their intended
validation scope. These inefficiencies further delay actionable feedback and consume GPU resources without proportionally
improving regression validation.

To address these issues, we present \sysname, a framework that optimizes both test selection and test
execution in CI pipelines for LLM training frameworks. For test selection, \sysname uses lightweight sampling to collect runtime traces from GPU-intensive tests and organizes them into a graph that records the functions and calls observed for each test.
Using this runtime evidence graph, \sysname selects only the tests that execute the changed functions and prunes those that execute them in equivalent contexts.
For test execution, \sysname schedules tests based on their execution history,
prioritizing high-risk tests to expose potential failures earlier, and accelerates individual tests by reducing
the workload along dimensions outside the intended validation scope of each test.

Evaluated on the CI workload of our LLM training framework, \sysname reduces the CI latency by 77.5\% and the GPU resource usage
by 63.9\% compared with path-based test selection, which is currently adopted in our CI pipelines. In addition, \sysname improves
the modified code coverage retention by 3.2\% over path-based test selection because runtime evidence recovers affected
tests missed by file-level mappings, thereby covering more modified code lines.
Furthermore, \sysname cuts the GPU resource usage before exposing the first failure by 89.2\%, compared with the currently used first-in-first-out
test scheduling. \sysname has now been integrated into the CI pipelines of our LLM training framework at \company. With \sysname handling
the CI workload, the average CI latency has been reduced by 58.4\%, substantially improving development efficiency
of our LLM training framework.

\section{Related Work}
\label{sec:related_work}

\textbf{Program analysis and runtime tracing for test selection.}
Regression test selection usually relies on dependencies between
code and tests to identify which tests are affected by code changes \citep{rothermel1996analyzing,yoo2012survey,elbaum2014techniques}.
In addition to manually maintained path mapping, static program
analysis is also widely used to extract these dependencies \citep{legunsen2017starts,zhang2018hyrts,liu2023precise,zhang2024hybridrts,maurina2025babelrts}.
File- or module-level static analysis is coarse-grained and still over-selects tests \citep{shi2017testoptimizer}.
Function- or method-level analysis can be more precise but introduce high analysis overhead for LLM training frameworks, causing the analysis to time out in our experiments.
To address this issue, some recent approaches optimize the static analysis process.
NameRTS makes function-level analysis potentially practical by using conservative name matching refined with runtime information
\citep{wang2026namerts}. However, end-to-end tests in LLM training frameworks often share a core training entry point but execute different
code paths across configurations, tasks, and enabled features. As NameRTS distinguishes tests by their test file entry points, it may therefore
select unnecessary jobs in this scenario, as evaluated in Section~\ref{sec:evaluation:effectiveness_of_test_selection}.

To extract precise dependencies between code and tests, some approaches turn to runtime tracing
\citep{gligoric2015ekstazi,gligoric2015dynamic,kauhanen2021pythonrts}. However, exhaustive tracing can substantially increase
the runtime and GPU resource consumption of CI pipelines \citep{wang2024ijacoco}. To address this issue, \sysname turns to lightweight
sampling to obtain runtime evidence, but it is challenging to design a test selection mechanism under potentially incomplete evidence.

\textbf{Prediction-based test selection and prioritization.}
Prediction-based approaches may leverage historical outcomes, developer-provided information, learned representations, or LLM reasoning
\citep{roy2026codesense,tao2025cgm} to
identify tests that are likely to fail \citep{leong2019assessing,machalica2019predictive,spieker2017reinforcement,lima2022coleman,pan2022slr,zhao2023revisiting,plyusnin2025targeted,mehta2021datadriven,herzig2015theo}.
However, their predictions do not provide a clear guarantee for test skipping. They are therefore
commonly used in fail-stop pipelines, where likely failures are executed first but the full test suite remains necessary if no failure is observed.
Rather than using predictions to skip tests, \sysname uses historical information only to schedule selected tests in our non-fail-stop CI pipelines.

\section{Overview}
\label{sec:overview}

\begin{figure}[t]
\centering
\vspace{-1em}
\includegraphics[width=0.85\textwidth]{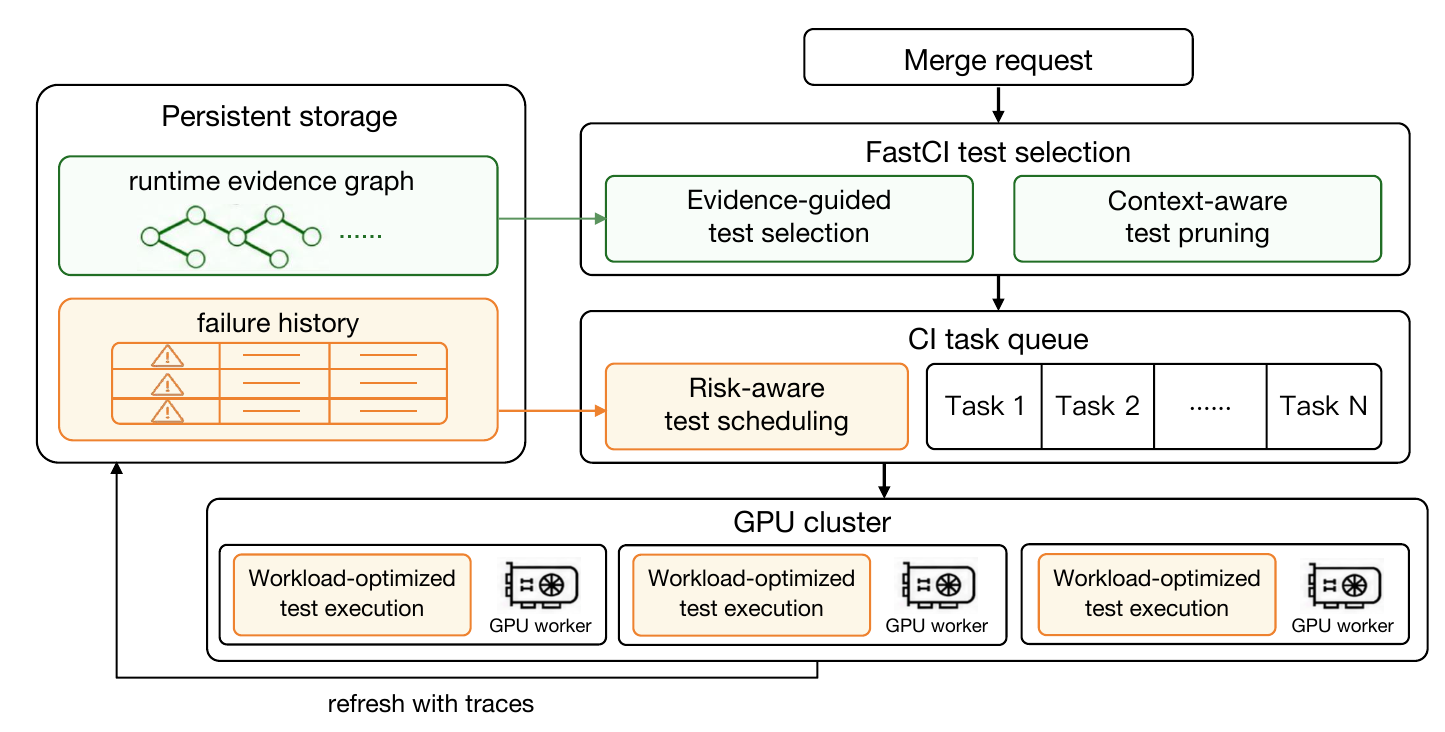}
\vspace{-0.5em}
\caption{Overview of \sysname.}
\label{fig:overview}
\vspace{-1.5em}
\end{figure}

Figure~\ref{fig:overview} provides an overview of \sysname. \sysname consists of modular components that can be integrated into existing CI pipelines
for LLM training frameworks with minimal changes. Across CI runs, it continuously updates a runtime evidence graph and test execution history. For each MR,
\sysname uses this state to optimize both test selection and execution, improving CI efficiency throughout the MR lifecycle.

Each CI run first enters test selection. \sysname applies \textbf{evidence-guided test selection}, mapping code changes to runtime graph entities and selecting tests
that reach them. If some changes lack corresponding entities, indicating incomplete runtime evidence, \sysname broadens the selection to reduce the risk of omitting affected tests.
The selected tests then pass through \textbf{context-aware test pruning}, which identifies tests that execute the changed code in
equivalent contexts and removes redundant ones from the selection. In this phase, the core of \sysname is a runtime evidence graph built from runtime-sampled traces.
This graph must be updated continuously to reflect the evolving codebase, which we will detail in Section~\ref{sec:design:evidence_guided_test_selection}.

The remaining tests then enter the test execution phase. \sysname first applies \textbf{risk-aware test scheduling} to prioritize high-risk tests
based on associations between code changes and test failures, together with the recent flakiness of each test. This scheduling exposes potential failures earlier,
allowing developers to receive feedback and intervene sooner. Then the ordered tests are submitted for execution, where \sysname applies
\textbf{workload-optimized test execution} to reduce workload along dimensions outside the intended validation scope of each test. If the workload cannot
be reduced, \sysname retains the original workload to preserve the validation capability of the test.

\section{Design}
\label{sec:design}

This section details the design of \sysname, consisting of four techniques, which are
evidence-guided test selection (Section~\ref{sec:design:evidence_guided_test_selection}), context-aware test pruning (Section~\ref{sec:design:context_aware_test_pruning}),
risk-aware test scheduling (Section~\ref{sec:design:risk_aware_test_scheduling}), and workload-optimized test execution (Section~\ref{sec:design:workload_optimized_test_execution}).

\subsection{Evidence-guided test selection}
\label{sec:design:evidence_guided_test_selection}

\begin{figure}[t]
\centering
\vspace{-1em}
\includegraphics[width=\textwidth]{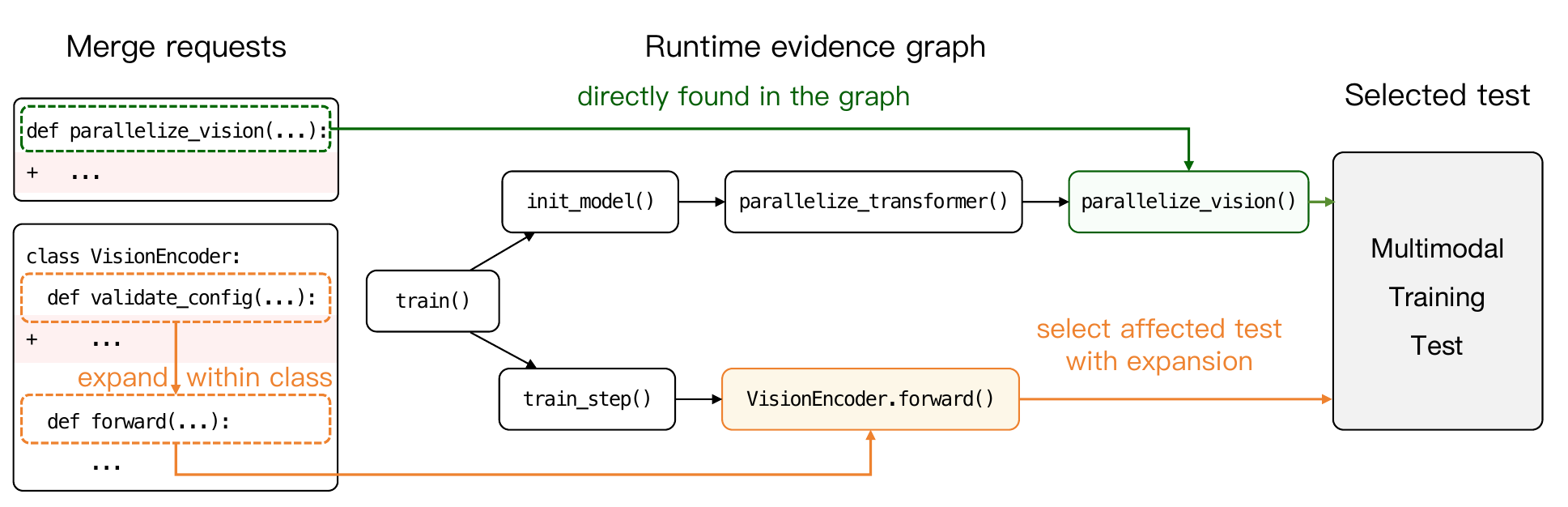}
\vspace{-1em}
\caption{Affected-test selection for two MRs, either directly from runtime evidence or through within-class expansion when evidence is missing.}
\label{fig:design:selection}
\vspace{-1.5em}
\end{figure}

Evidence-guided test selection uses a runtime evidence graph built from runtime traces sampled during previous CI runs. When the
graph contains evidence for a modified function, \sysname directly selects the tests observed to execute that function. When such
evidence is missing, \sysname intentionally broadens the selection to include tests that may execute the modified function even though
this execution was not observed in previous runs.

For each codebase version, \sysname maintains a runtime evidence graph $G = (V, E, A_V, A_E)$ that captures function-level code--test dependencies.
$V$ contains observed functions and methods, while $E \subseteq V \times V$ represents observed calls. Attributes $A_V$ and $A_E$ map vertices and edges to the tests
that executed them. Node attributes support precise affected-test selection, while edges support context-aware pruning in Section~\ref{sec:design:context_aware_test_pruning}.

At the beginning of the test selection, code changes are mapped to a set of corresponding entities, denoted by $M$. For changes within explicit
function or method scope, $M$ contains the corresponding function or method. For changes outside explicit function or method scope,
such as global variable definitions and conditional imports, $M$ includes all functions and methods within the changed file.

Then \sysname searches the runtime evidence graph $G$ for each entity in $M$ and extracts the tests with explicit evidence
of executing the modified code. However, it is possible that some entities in $M$ cannot be found in $G$, indicating that the runtime
evidence is incomplete. This issue stems from the sampling method for trace collection, especially for short-lived functions like
configuration setters and utility functions. To reduce the risk of omitting potentially affected tests, \sysname intentionally broadens
the test selection in two ways. First, for each modified method, $M$ is expanded to include all methods within the same class, as
multiple methods of a class are usually invoked together within the same test. Second, for each modified function, $M$ is expanded
to include all its caller functions by scanning the codebase. After the expansion, evidence associated with these added entities
allows \sysname to identify candidate tests for modified entities missing from $G$.
Let $Extend(M)$ denote the expanded set of entities. The selected tests $T_{affected}(M)$ can be determined as
\begin{equation}
T_{affected}(M) = \bigcup_{m \in Extend(M)} A_V(m)
\end{equation}

Figure~\ref{fig:design:selection} illustrates the test selection process. For an MR modifying \texttt{parallelize\_vision},
its corresponding entity can be found in the runtime evidence graph and the affected test is selected directly. However, for an MR modifying
\texttt{VisionEncoder.validate\_config}, which is a simple configuration setter, its corresponding entity is missing from the graph.
Therefore, \sysname broadens the selection through other methods within the same class and finally identifies the test through \texttt{VisionEncoder.forward}.
This test was verified to be affected by the MR.

The remaining challenge is keeping the runtime evidence graph aligned with the evolving codebase. Sampling tools (\emph{e.g.}, \texttt{py-spy}
\citep{frederickson2026pyspy})
add overhead, but it is small relative to savings of \sysname. By default, \sysname samples each CI run and updates the graph when an MR
is merged. For a stable codebase, since a single MR changes only a small fraction of the codebase, the graph tolerates limited staleness and can instead be updated periodically,
such as every three commits.

\subsection{Context-aware test pruning}
\label{sec:design:context_aware_test_pruning}

After evidence-guided test selection, the selected test set may still be large because multiple tests can execute the same modified
function \citep{shi2015combining}. For instance, lightweight environment checking functions are invoked by most end-to-end tests, causing many tests with
redundant coverage to be selected when these functions change. Although one representative test may be sufficient, tests cannot be
pruned solely because they execute the same function, as the behavior of the function may depend on the calling context. \sysname
therefore distinguishes tests by their execution contexts and prunes only those that exercise the modified function in equivalent contexts.

To identify such redundant tests, \sysname extracts an execution context $C_K(m,t)$ for each modified entity $m$ and selected test $t$
from the runtime evidence graph. The context is a test-specific subgraph of $G$ containing the callers and callees within $K$ hops of $m$.
We use $K=3$ by default.
Then two tests are considered redundant for $m$ only if their context subgraphs are identical and their validation metadata, such as task type
and hardware requirements, also match. Accordingly, test equivalence is defined as follows:
\begin{equation}
t_i \equiv_m t_j \iff C_K(m,t_i) = C_K(m,t_j) \land meta(t_i) = meta(t_j).
\end{equation}
Pruning is performed independently for each modified entity, and the retained tests are then merged across all modified entities.

\begin{figure}[t]
\centering
\vspace{-1em}
\includegraphics[width=\textwidth]{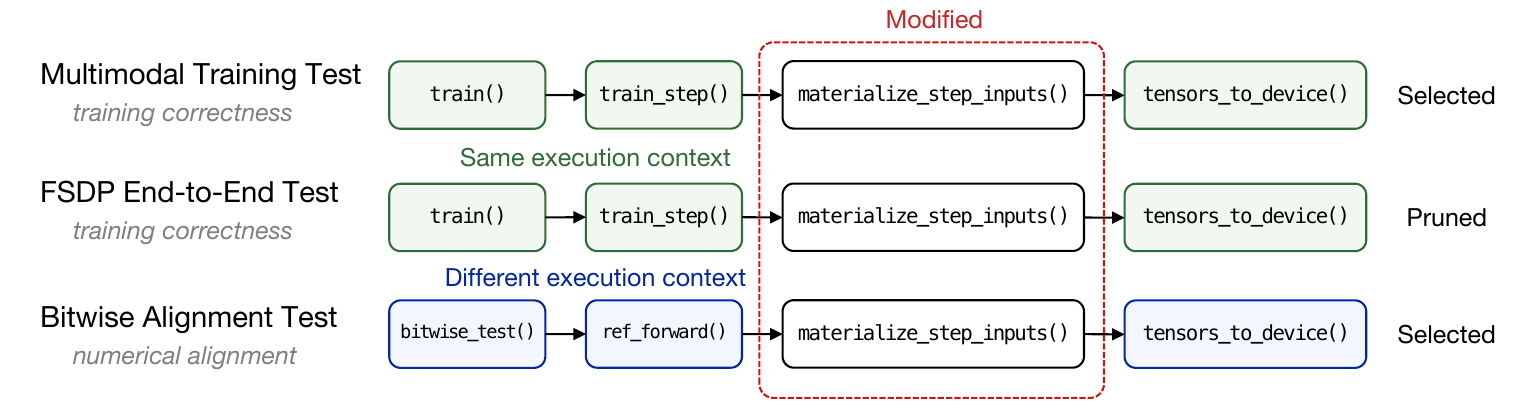}
\vspace{-2em}
\caption{Context-aware test pruning when \texttt{materialize\_step\_inputs} is modified.}
\label{fig:design:pruning}
\vspace{-1em}
\end{figure}

Figure~\ref{fig:design:pruning} shows an example of redundant tests. The Multimodal Training Test and FSDP End-to-End Test execute
\texttt{materialize\_step\_inputs} with identical caller and callee chains in the runtime evidence graph. They are therefore considered redundant,
and only one needs to be retained. In contrast, the Bitwise Alignment Test executes the function in a different context and must
remain selected.

After identifying redundant tests, \sysname selects a representative test for each equivalence group.
It estimates the GPU cost of each test from its historical execution time and the number of GPUs required. Among tests with equivalent execution contexts and validation metadata, \sysname selects the test with the lowest
estimated GPU cost, while the remaining tests are pruned from the selection.

\subsection{Risk-aware test scheduling}
\label{sec:design:risk_aware_test_scheduling}

Once the test set has been selected and pruned, the remaining tests are scheduled and dispatched to the GPU resource pool for execution. First-in-first-out scheduling can delay actionable failure feedback.
Developers must then fix the problem and rerun CI from scratch, leading to inefficient development cycles and GPU resource waste. \sysname instead derives each test's risk from historical execution
outcomes and prioritizes high-risk tests to expose potential failures earlier.

Failures in production CI pipelines arise not only from incorrect code changes but also from external factors such as environment issues
and dependency breakages. \sysname therefore uses two complementary risk indicators to prioritize tests. The first captures historical associations
between code changes and test failures. It identifies failures likely caused by code changes but provides limited evidence for newly added
modules and functions. The second captures the recent instability of each test. It reflects failures caused by changing environments, dependencies,
or flaky behavior \citep{luo2014empirical,bell2018deflaker,lam2020lifecycle}, but is less specific to the current code change.

\sysname assesses the risk of each test by combining the two risk indicators as
\begin{equation}
    R(t) = \alpha_t R_{code}(t) + (1 - \alpha_t) \times R_{instability}(t)
    \label{eq:risk_score}
\end{equation}
where $R_{code}(t)$ is the risk of $t$ based on code changes, and $R_{instability}(t)$ is the risk of $t$ based on recent instability.
$\alpha_t$ reflects the confidence in the code change risk indicator and increases with the number of historical observations available for the
current change-test associations. Appendix~\ref{sec:appendix:risk_score_computation} details the calculation of this risk score.
By combining these two indicators, \sysname can capture both failures caused by code changes and external
factors, and avoid being misled when encountering newly added modules and functions. Accordingly, \sysname schedules tests in descending order
of their risk scores, improving the feedback efficiency and reducing the GPU resource waste caused by running on already broken code.

\subsection{Workload-optimized test execution}
\label{sec:design:workload_optimized_test_execution}

Beyond test scheduling, \sysname also considers the workload of each test to optimize the GPU resource utilization. Typically, the
workload of each test is derived from production configurations. Some dimensions of the workload, including parallelism degree and
batch size, are adjusted to CI requirements, but others like model size, sequence length and production options (\emph{e.g.}, logging and metrics
collection) are left unchanged. These unchanged dimensions of the workload further increase GPU resource usage without proportionally improving
regression validation, as they are outside the intended validation scope of the test.

For each test, \sysname identifies the dimensions of the workload that are related to its intended validation scope through test metadata
and its historical execution paths from the runtime evidence graph. For example, tests for distributed execution rely on the parallelism degree,
while tests for long sequence training rely on the sequence length. For dimensions that can be explicitly identified as unrelated to the
intended validation scope, \sysname reduces them automatically. For dimensions that cannot be explicitly identified, \sysname leaves them unchanged.

\sysname applies these reductions as configuration overrides when submitting a test to the existing CI runner, without modifying the test code.
In practice, we found that decreasing model size and turning off production options like logging and metrics collection can accelerate tests
in many cases while producing the same test outcomes as the original workload in our comparisons. To reduce the risk of missing failures that appear only under the original workload,
\sysname periodically executes the same test with both the original and reduced workloads and compares their pass or fail outcomes. The reduction is retained only while the outcomes agree.
Otherwise, the original workload is restored. Repeating this comparison allows \sysname to withdraw an accepted reduction when the code or test behavior changes.

\section{Evaluation}
\label{sec:evaluation}

We evaluate \sysname on CI workloads from our LLM training framework, which supports diverse training tasks at \company. Compared with deployed CI, \sysname reduces
latency by 77.5\% and GPU usage by 63.9\%, while improving modified code coverage retention by 3.2\% through recovering affected tests missed by file-level mappings.
Each technique contributes to overall performance. \sysname has been integrated into production CI at \company and now handles the CI workload, reducing CI latency by 58.4\%, with no reported regressions caused by omitted tests across 784 CI runs over three weeks.

\subsection{End-to-End CI Performance}
\label{sec:evaluation:end_to_end_ci_performance}

\begin{figure}[t]
\vspace{-0.5em}
\centering
\begin{subfigure}[b]{0.24\textwidth}
    \centering
    \includegraphics[width=\linewidth]{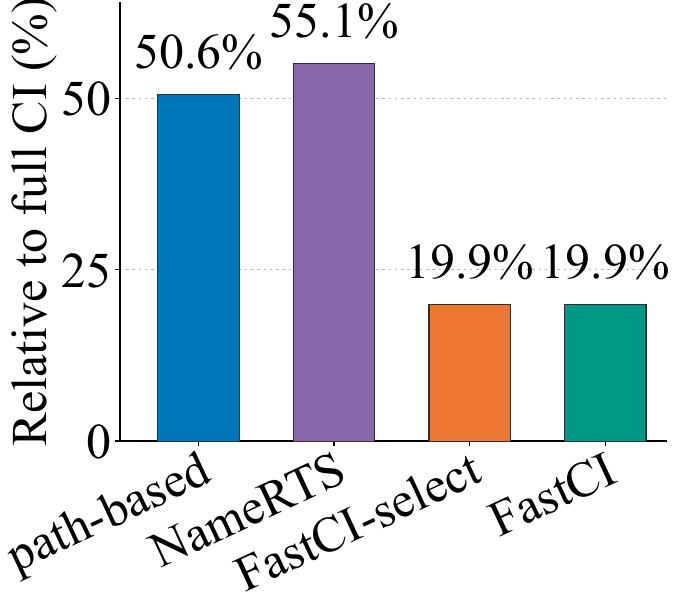}
    \caption{\# selected tests.}
\end{subfigure}
\hfill
\begin{subfigure}[b]{0.24\textwidth}
    \centering
    \includegraphics[width=\linewidth]{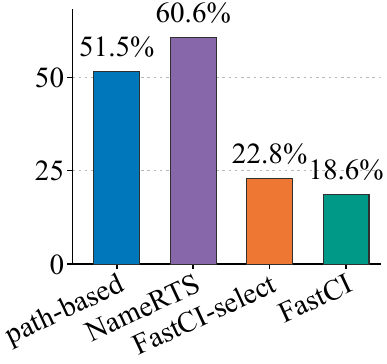}
    \caption{GPU resource usage.}
\end{subfigure}
\hfill
\begin{subfigure}[b]{0.24\textwidth}
    \centering
    \includegraphics[width=\linewidth]{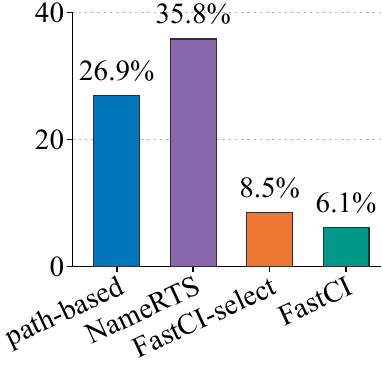}
    \caption{CI latency.}
\end{subfigure}
\hfill
\begin{subfigure}[b]{0.24\textwidth}
    \centering
    \includegraphics[width=\linewidth]{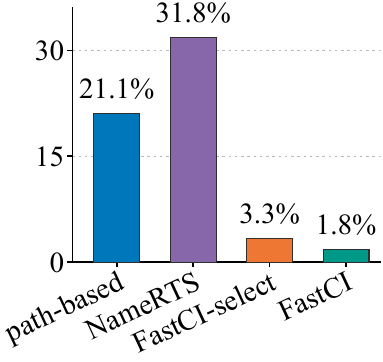}
    \caption{Queueing time.}
\end{subfigure}
\vspace{-0.5em}
\caption{End-to-end CI performance. The statistics are relative to full CI.}
\vspace{-0.5em}
\label{fig:eval:performance}
\end{figure}

\begin{figure}[t]
\vspace{0em}
\centering
\begin{subfigure}[b]{0.492\textwidth}
    \centering
    \includegraphics[width=\linewidth]{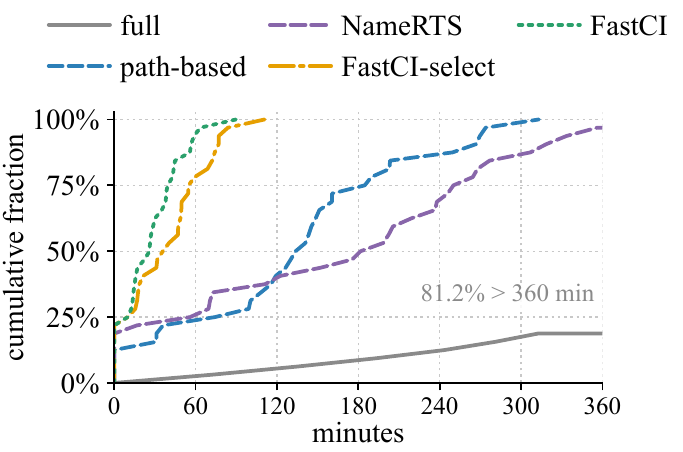}
    \caption{CI latency.}
\end{subfigure}
\hfill
\begin{subfigure}[b]{0.492\textwidth}
    \centering
    \includegraphics[width=\linewidth]{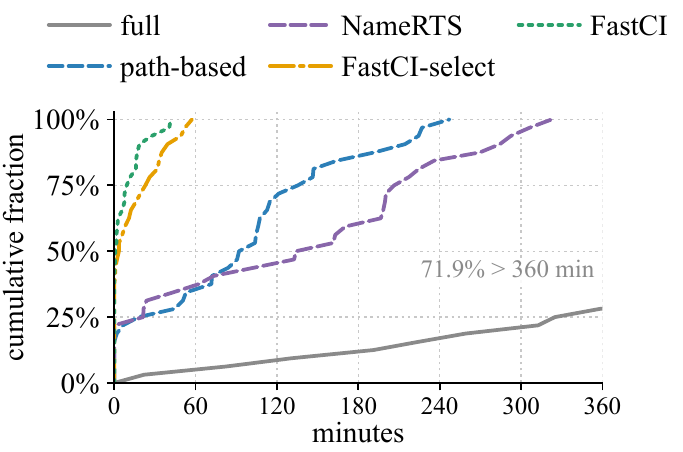}
    \caption{Queueing time.}
\end{subfigure}
\vspace{-0.5em}
\caption{CDF of CI latency and queueing time under different approaches. The x-axis is shown up to 360 minutes and annotations report the fraction of full CI runs beyond this range.}
\label{fig:eval:performance_cdf}
\vspace{-1.5em}
\end{figure}

We evaluate the end-to-end performance of \sysname by sampling production MRs and submitting them to self-provisioned shadow CI pipelines over two days, avoiding interference with production CI.
We compare \sysname against full CI, path-based CI, NameRTS, and \sysname-select. Under full CI, the entire suite is executed for each MR, while path-based CI uses the file-level
mapping to select tests and is currently deployed in our production CI. NameRTS is a state-of-the-art regression test selection method whose optimized static analysis scales to our large LLM training framework~\citep{wang2026namerts}.
\sysname-select applies only evidence-guided selection and context-aware pruning, without risk-aware test scheduling or workload-optimized execution.

Figure~\ref{fig:eval:performance} compares the average efficiency of different approaches. Path-based CI selects 50.6\% of tests and reduces GPU resource usage, CI latency,
and queueing time of tests to 51.5\%, 26.9\%, and 21.1\% of full CI, respectively. However, its coarse-grained mappings still select many unnecessary tests,
preventing it from fully relieving the pressure on production CI. Fine-grained static analysis alone is also insufficient as NameRTS selects 55.1\% of tests and is even less efficient
than path-based CI across all four metrics. This is because NameRTS is only able to map code changes to entry points of tests, which can be shared by the end-to-end tests of LLM training framework CI, causing over-selection. 
Beyond them, \sysname reduces GPU resource usage, CI latency, and queueing time
to 18.6\%, 6.1\%, and 1.8\% of full CI.
% Compared with path-based CI, \sysname can reduce GPU resource usage by 63.9\%, CI latency by 77.5\%, and queueing time of tests by 91.6\%.

Figure~\ref{fig:eval:performance_cdf} shows CI latency and test queueing time distributions under different approaches. For both metrics,
\sysname outperforms path-based CI and NameRTS at every percentile. Compared with path-based CI, at the median, \sysname reduces CI latency by 79.2\%, and queueing time by 97.3\%.
At the 90th percentile, the corresponding reductions are 78.3\% and 89.0\%, demonstrating that the improvement is consistent across MRs rather than being
driven by a few inexpensive cases. Notably, some MRs have zero CI latency under path-based CI, NameRTS, and \sysname. They usually
introduce new features or new modules and thus there are no tests executing the newly added code.

\subsection{Effectiveness of Test Selection}
\label{sec:evaluation:effectiveness_of_test_selection}

\begin{table}[t]
\vspace{0em}
\caption{Comparison between different test selection methods.}
\label{tab:eval:selection}
\centering
\renewcommand{\arraystretch}{1.12}
\begin{tabular}{lcccc}
\toprule
Method & \makecell{Number of \\ selected tests} & \makecell{GPU resource \\ cost} & \makecell{Recall of \\ affected tests} & \makecell{Modified code \\ coverage retention} \\
\midrule
Path-based & 62.6\% & 67.6\% & 76.1\% & 96.0\% \\
NameRTS & 77.2\% & 80.8\% & 94.0\% & 95.8\% \\
LLM reasoning & 39.4\% & 42.1\% & 54.6\% & 83.1\% \\
\sysname-select w/o pruning & 35.9\% & 39.4\% & 84.5\% & 99.8\% \\
\sysname-select & 25.7\% & 28.6\% & 66.2\% & 99.2\% \\
\bottomrule
\end{tabular}
\vspace{-1.5em}
\end{table}

We evaluate test selection by replaying production CI workloads from an internal LLM training framework.
We compare five test selection methods: path-based test selection, LLM reasoning, NameRTS, \sysname-select without context-aware test pruning, and the full \sysname-select.
For the LLM reasoning baseline, a GPT-5.5~\citep{openai2026gpt55} agent analyzes each MR diff and candidate tests and returns those it considers affected.
For efficiency, we measure the selected-test ratio and normalized GPU cost. For coverage, we execute the full suite with exhaustive line-level collection using
\texttt{Coverage.py} \citep{batchelder2026coveragepy}, establishing an offline reference of tests that execute the modified code. Against this reference, we measure
the recall of affected tests and the modified code coverage retained by each selection method.

As shown in Table~\ref{tab:eval:selection}, compared with path-based test selection, evidence-guided test selection achieves better efficiency by reducing the selected
test ratio from 62.6\% to 35.9\% and normalized GPU cost from 67.6\% to 39.4\%, while improving the recall of affected tests from 76.1\% to 84.5\% and modified code coverage
retention from 96.0\% to 99.8\%.
By further applying context-aware test pruning, \sysname-select reduces the selected test ratio to 25.7\% and normalized GPU cost to 28.6\%. This efficiency gain comes
with recall of affected tests decreasing to 66.2\%, as
some affected tests are intentionally pruned as redundant. 
However, modified code coverage
retention decreases by only 0.6\% to 99.2\%, which remains higher than that of the other approaches, suggesting that most pruned tests cover modified code already
executed by the retained tests.

As for other baselines, LLM reasoning reduces the selected test ratio and GPU cost compared to path-based test selection, but it performs poorly on coverage,
retaining only 54.6\% of affected tests and 83.1\% of modified code coverage.
This is because LLM reasoning cannot infer the complete execution paths of end-to-end tests precisely, especially on our large codebase.
For NameRTS, although it achieves 94.0\% recall of affected tests, its normalized GPU cost also reaches 80.8\%, indicating that the higher recall substantially comes from over-selection.

\subsection{Effectiveness of Test Scheduling}
\label{sec:evaluation:effectiveness_of_test_scheduling}

We evaluate whether \sysname's risk-aware scheduling exposes CI failures earlier, improving feedback efficiency and reducing GPU usage. We compare \sysname with
the production first-in-first-out order, the two single-indicator variants, and COLEMAN,
a representative learning-based test prioritization approach for CI~\citep{lima2022coleman}. We use its best-performing configuration, FRRMAB with the TimeRank reward,
and average the results over ten fixed random seeds.
Because resource pool capacity and concurrent CI workloads affect feedback latency, we focus on first-failing-test rank and GPU consumption before the first failure,
which directly reflect the effectiveness of test scheduling.

\begin{figure}[H]
\vspace{-0.5em}
\centering
\includegraphics[width=\textwidth]{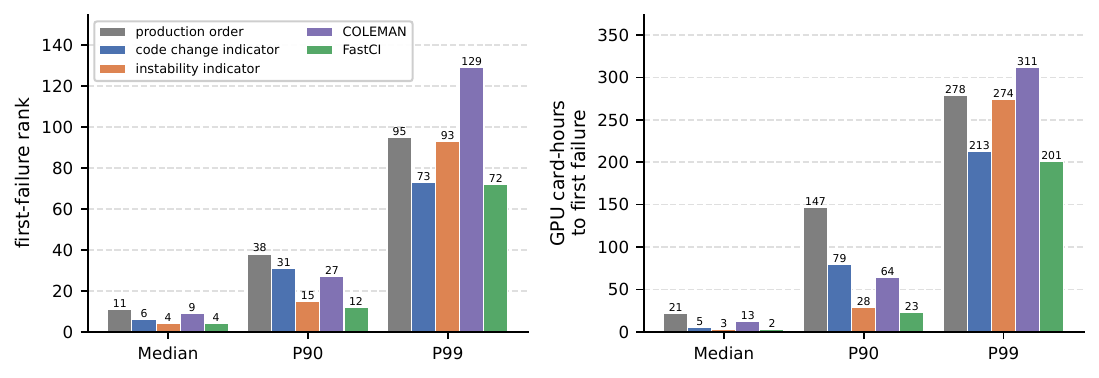}
\vspace{-2em}
\caption{Effectiveness of risk-aware test scheduling compared with production order, single-indicator variants, and COLEMAN.}
\label{fig:eval:scheduling}
\vspace{-1.5em}
\end{figure}

Figure~\ref{fig:eval:scheduling} shows that, compared with production first-in-first-out scheduling, \sysname reduces median first-failure rank by 63.6\% and median GPU
consumption before the first failure by 89.2\%, exposing failures earlier with substantially less GPU consumption. Compared with COLEMAN, \sysname reduces the median
first-failure rank from 9 to 4 and the median GPU consumption before the first failure from 12.6 to 2.3 card-hours. It also performs better at P90 and P99 on both metrics.
The two indicators provide complementary benefits.
The instability indicator performs better at the median and P90,
suggesting that recent test instability is more informative for most production failures. In contrast, the code change indicator performs better at P99, where CI runs tend to contain larger
selected test sets. By combining both indicators, \sysname achieves better overall performance than either indicator alone.

\subsection{Effectiveness of Workload Optimization}
\label{sec:evaluation:effectiveness_of_workload_optimization}

We evaluate two questions: whether workload-optimized execution accelerates individual tests and whether this acceleration improves end-to-end CI.
Comparing tests with and without workload optimization, we find that \sysname reduces average single-test execution time by 11.8\% and GPU consumption by 19.3\%,
since tests with larger workloads tend to run longer and therefore offer more optimization opportunities. This acceleration of individual tests contributes to the
end-to-end CI performance. As illustrated in Figure~\ref{fig:eval:performance}, compared with \sysname-select, \sysname can further reduce the GPU resource usage by 18.3\%,
average CI latency by 28.6\% and queueing time of tests by 46.6\%, demonstrating the effectiveness of workload-optimized test execution in an end-to-end setting.

\subsection{Overhead and Coverage of Runtime Sampling}
\label{sec:evaluation:overhead_and_coverage_of_runtime_sampling}

\begin{table}[t]
\vspace{0em}
\caption{Tracing overhead and final test-selection effectiveness at different sampling rates.}
\vspace{-0.5em}
\label{tab:eval:sampling}
\centering
\renewcommand{\arraystretch}{1.12}
\begin{tabular}{lccc}
\toprule
Sampling rate & \makecell{Tracing overhead}  & \makecell{Recall of \\ affected tests} & \makecell{Modified code \\ coverage retention} \\
\midrule
10~Hz & 0.3\% & 57.7\% & 97.5\% \\
50~Hz & 4.1\% & 66.2\% & 99.2\% \\
100~Hz & 17.0\% & 68.4\% & 99.6\% \\
\bottomrule
\end{tabular}
\vspace{-1.5em}
\end{table}

We study the sampling-rate tradeoff and whether runtime sampling overhead becomes a new CI bottleneck. We repeat the test selection evaluation in
Section~\ref{sec:evaluation:effectiveness_of_test_selection} at 10, 50, and 100~Hz, enabling context-aware pruning in all configurations. Thus, the reported recall and coverage reflect
final selection rather than raw trace capture. In this evaluation, Hz denotes stack samples collected per second.
Table~\ref{tab:eval:sampling} shows that higher rates provide more complete runtime evidence for coverage but add overhead. In practice, 50~Hz is a suitable tradeoff,
retaining 99.2\% of full-suite modified-code coverage with only 4.1\% overhead. Raising the rate to 100~Hz improves retention by only 0.4\% but raises overhead
to 17.0\%. Relative to \sysname's efficiency gains, this overhead is small and does not become a new CI bottleneck, as the end-to-end evaluation
(Section~\ref{sec:evaluation:end_to_end_ci_performance}) still shows substantial reductions in CI latency and GPU usage.

\section{Conclusion}
\label{sec:conclusion}

We present \sysname, a framework that improves CI efficiency for LLM training frameworks by using a runtime evidence graph to select and prune tests while optimizing
their scheduling and execution. On production CI workloads, \sysname reduces CI latency by 77.5\% and GPU resource usage by 63.9\%, while improving modified code coverage
retention by 3.2\% over the deployed CI pipelines. \sysname has been integrated into production CI at \company.

\subsection*{AI Use Statement}

Generative AI tools were used to assist with language editing, literature search
and reference organization, and the implementation of plotting scripts. The
manuscript was originally written by the authors, and all AI-assisted textual
suggestions were reviewed and manually revised by the authors. The conceptual
design and content of all figures were determined by the authors. For
experimental figures, the authors provided the data, verified the plotting code,
and executed the scripts themselves. Generative AI was not used to generate
experimental data, conduct experiments, formulate the core methodology, or draw
final scientific conclusions. The authors take full responsibility for the final
content of this work.

\bibliography{iclr2027_conference}
\bibliographystyle{iclr2027_conference}

\clearpage
\appendix
\section{Risk Score Computation for Test Scheduling}
\label{sec:appendix:risk_score_computation}

We calculate the risk score in Equation~\ref{eq:risk_score} from statistics
collected online from previous CI runs. In the evaluation, code changes are
represented by the files containing the modified entities. For a changed file
$f$ and test $t$, let $n_{f,t}$ be the number of previous CI runs in which $f$
was changed and $t$ was executed, and let $e_{f,t}$ be the number of these runs
in which $t$ failed. Given the set of files $F$ changed by the current MR, risk
indicated by code change is
\begin{equation}
R_{code}(t) = \max_{f \in F} \frac{e_{f,t}}{n_{f,t}},
\end{equation}
where a pair without previous observations contributes zero. Let $f^*$ be the
file attaining the maximum positive failure rate. We set the confidence of this
signal to
\begin{equation}
\alpha_t = \frac{n_{f^*,t}}{n_{f^*,t}+\lambda}.
\end{equation}
Here, $\lambda$ is a smoothing parameter controlling how quickly the code change
indicator gains confidence; we set $\lambda=5$ in our evaluation. If no changed
file has a positive historical failure rate for $t$, we set $R_{code}(t)=0$ and
$\alpha_t=0$, so the risk score is determined solely by the instability
indicator.

To assess test instability, we retain up to $W$ most recent execution outcomes
for each test. Let $N_t$ be the number of outcomes available for test $t$ and
$L_t=\min(W,N_t)$ be the actual history length used. Let $y_i \in \{0,1\}$
denote the $i$-th outcome in this history, where one indicates failure and the
outcomes are ordered from oldest to newest. We first compute an exponentially
weighted failure rate
\begin{equation}
\widetilde{R}_{instability}(t) =
\frac{\sum_{i=1}^{L_t} \gamma^{L_t-i} y_i}{\sum_{i=1}^{L_t} \gamma^{L_t-i}},
\end{equation}
where $\gamma$ is the exponential decay factor. We set $W=5$ and $\gamma=0.8$
in our evaluation, and the score is zero when no history is available.
Consecutive passes may indicate that a previous failure has been fixed or that
test optimizations have reduced its flakiness. \sysname therefore increasingly
attenuates the exponentially weighted failure rate with the consecutive-pass
count, which resets after a failure. In our evaluation, the attenuation factors
are $0.85$ and $0.6$ at two and four consecutive passes, respectively, and the
attenuated value is used as $R_{instability}(t)$.

Tests are ordered by decreasing $R(t)$. Ties preserve the original execution
order. To avoid future-information leakage in the offline replay, MRs are
processed chronologically, both statistics are updated only after the current MR
has been scored, and the first 30 MRs are used only to warm up the history.

\section{Maintaining the Runtime Evidence Graph}
\label{sec:appendix:runtime_evidence}

\sysname injects lightweight runtime sampling into the existing CI launch script
rather than modifying individual tests. During each test, \texttt{py-spy} samples
the rank-0 Python process at the configured rate and records sampled stacks in
the Speedscope format. Subprocess following can also be enabled. Because sampling
may miss short-lived functions and native execution, an entity absent from the
trace is treated as missing evidence, activating the expansion mechanism in
Section~\ref{sec:design:evidence_guided_test_selection}.

Trace files are organized by test, with each test corresponding to one CI job in
our deployment. For each sampled stack, frames outside the repository are
removed, paths are normalized, and Python methods are resolved to class-qualified
names using the source AST when possible. Consecutive retained frames produce
caller--callee edges, while each frame records the test that executed the
function. All trace files for a test are combined into one partial graph.

Before selecting tests for an MR, \sysname loads the graph associated with its
merge base, falling back to the stable graph if that version is unavailable, and
saves it as the base snapshot for the current run. Each successfully traced test
produces a partial graph. After the CI run, \sysname replaces the base snapshot's
evidence for these tests with their partial graphs, preserves evidence for tests
not executed in the run, and stores the merged graph under the new commit. A
later MR whose merge base matches that commit uses the updated graph for
selection, and its execution produces the next set of traces, closing the
collection--selection--update loop.

\section{Evaluation Details}
\label{sec:appendix:experimental_setup}

\subsection{LLM Selection Baseline}
The LLM selection baseline operates on the same MR revisions and candidate tests
as the other methods. We implement it with a GPT-5.5 code agent given the MR diff,
the evaluated repository, and the CI test definitions. Its fixed instruction asks
the agent to interpret the semantic effect of the changes, inspect related code
and concrete CI entry points, and select candidate tests whose execution may
reach the changed behavior. When the impact is uncertain, the agent favors recall
while keeping its output within the MR-triggered candidate test set. We use the
same instruction and repository-access scope for every evaluated MR. No execution
outcome from the current or future MRs is provided.

\subsection{COLEMAN Scheduling Baseline}
To isolate the effect of test ordering, all schedulers receive the same selected
tests, per-test runtime and GPU requirements, and MR order. COLEMAN treats each
CI test as an arm and uses its FRRMAB scheduler with the TimeRank reward. Before
each MR, FRRMAB combines the normalized TimeRank reward accumulated over a
sliding window with an exploration bonus for less frequently observed tests.
After the resulting order is replayed, TimeRank converts the test outcomes and
their positions into rewards that favor placing failures earlier. We configure
FRRMAB with $C=0.3$, a history window of 100 CI cycles, and a decay factor of 1,
following the original evaluation~\citep{lima2022coleman}. Equal-score ties are
broken using a seeded random order, and we average the results over ten fixed
seeds. For each MR, COLEMAN produces the ranking using only outcomes from earlier
MRs. The current MR's outcomes are added to its history only after ranking and
replay, preventing future-information leakage.

\subsection{Offline Coverage Reference}
The evaluation executes the complete CI test suite and collects per-job line
coverage using \texttt{Coverage.py}. Coverage fragments within each job are
combined before its XML report is exported. We retain
positive-hit lines, normalize paths against the evaluated repository revision,
and map them to executable lines inside Python function bodies, excluding
module-level code, declarations, decorators, docstrings, comments, and blank
lines. A test enters the offline affected-test reference for a modified function
only if its report shows that execution entered the function body. Modified-code
coverage retention is the fraction of executable lines in modified functions
covered by the full CI suite that are also covered by the selected tests. A
retention of 100\% means that the selected tests execute every modified Python
line covered by the full CI suite, but tests covering the same lines may exercise
different inputs and assertions. The reference also excludes native C++ and CUDA
execution.

\section{Production Deployment}
\label{sec:appendix:deployment}

\subsection{Deployment Process}
We first implemented the complete FastCI workflow and exercised it end to end on
the self-provisioned shadow CI used in our evaluation. This environment allowed
us to validate the workflow and collect the experimental evidence without
affecting production CI. We then reviewed the measured improvements and observed
failure modes with the development team before enabling FastCI for production
workload.

We rolled out FastCI in four stages. We first enabled runtime tracing for a small
subset of CI tests to validate graph construction and maintenance. We then
introduced an MR-title keyword that routed selected MRs through the complete
FastCI workflow. Next, we expanded the set of instrumented tests to validate
graph construction across the CI suite. After these checks, we removed the
keyword gate and enabled FastCI by default for eligible production MRs.

\subsection{Deployment Safeguards}
We maintained compatibility with the existing CI workflow throughout the rollout.
For non-code configuration changes, such as YAML only updates, the test selection
stage continued to use the existing CI rules rather than FastCI's evidence guided
test selection. The staged rollout allowed us to expand FastCI's scope incrementally
and validate each stage before proceeding.

To complement the staged rollout, we implemented a configuration-based rollback
mechanism. A single configuration field supports two rollback levels: FastCI can
stop handling incoming MRs, returning them to the existing CI workflow, or be
disabled entirely. Both levels require only a configuration update, enabling
rapid containment without code changes.

\subsection{Operational Lessons}

\paragraph{On-demand graph materialization.}
During early validation, only a small number of traces had accumulated, and
graph maintenance appeared inexpensive. Our initial implementation eagerly
processed all accumulated traces and maintained graphs for every version. In
production, however, the amount of trace processing performed for each MR grew
as traces accumulated, eventually causing the test-selection step to time out.
We therefore switched to on-demand graph materialization. For each MR, FastCI
processes only the traces associated with the version whose graph is required.
This makes the trace-processing overhead per MR $O(1)$ with respect to the total
number of accumulated traces.

\paragraph{Local configuration derivation.}
FastCI requires a small set of test-specific settings for each CI test run. Our
initial design centrally managed these settings in persistent storage, requiring
each test to perform remote reads and writes during execution. Across many
concurrently running tests, this remote I/O introduced considerable overhead. We
therefore pass the necessary inputs through environment variables and let each
test derive its settings locally. This removes remote storage access from the
test execution path and nearly eliminates the configuration overhead.

\end{document}